\documentclass{article}

\usepackage{arxiv}
\usepackage{amsmath}
\usepackage[utf8]{inputenc} 
\usepackage[T1]{fontenc}    
\usepackage{hyperref}       
\usepackage{url}            
\usepackage{booktabs}       
\usepackage{amsfonts}       
\usepackage{nicefrac}       
\usepackage{microtype}      
\usepackage{lipsum}
\usepackage{mathrsfs}
\usepackage{graphicx}

\usepackage[utf8]{inputenc}
\usepackage{algorithm}
\usepackage{algpseudocode}

\usepackage{amsfonts}
\usepackage{float}

\usepackage{amssymb}  
\usepackage{graphicx}
\usepackage{dcolumn}
\usepackage{bm}
\usepackage{nccmath}

\usepackage{algpseudocode}
\usepackage{verbatim}
\usepackage[table]{xcolor}
\usepackage{tabularx}
\graphicspath{ {./images/} }

\title{Energy-Guided Data Sampling for Traffic Prediction with Mini Training Datasets}

\author{
 Zhaohui Yang \\
  \texttt{zhaohuiyang@google.com} \\
   \And
 Kshitij Jerath \\
  \texttt{Kshitij\_Jerath@uml.edu} \\
}

\begin{document}
\maketitle
\begin{abstract}
Recent endeavors aimed at forecasting future traffic flow states through deep learning encounter various challenges and yield diverse outcomes. A notable obstacle arises from the substantial data requirements of deep learning models, a resource often scarce in traffic flow systems. Despite the abundance of domain knowledge concerning traffic flow dynamics, prevailing deep learning methodologies frequently fail to fully exploit it. To address these issues, we propose an innovative solution that merges Convolutional Neural Networks (CNNs) with Long Short-Term Memory (LSTM) architecture to enhance the prediction of traffic flow dynamics. A key revelation of our research is the feasibility of sampling training data for large traffic systems from simulations conducted on smaller traffic systems. This insight suggests the potential for referencing a macroscopic-level distribution to inform the sampling of microscopic data. Such sampling is facilitated by the observed scale invariance in the normalized energy distribution of the statistical mechanics model, thereby streamlining the data generation process for large-scale traffic systems. Our simulations demonstrate promising agreement between predicted and actual traffic flow dynamics, underscoring the efficacy of our proposed approach.

\end{abstract}


\section{Introduction}


The ability to accurately and efficiently estimate future traffic states is a key component of controlling traffic flow and mitigating congestion, which is an important goal for the transportation research community. Fortunately, the combination of increased vehicular connectivity (which can generate real-time traffic condition updates), availability of increased computational power (which can process the incoming data streams), and progress in machine learning algorithms (which can extract knowledge patterns from the processed data), has resulted in the unique opportunity to make potentially usable predictions about future states of traffic flow, including congestion. However, the prediction of emergent dynamics in large-scale, self-organizing systems (such as congestion patterns in traffic flow) remains a challenging task, and significant work needs to be performed to achieve this goal \cite{Shalizi2001}\cite{Shalizi2004}.


Some prior works have relied on data-driven methods to predict future behaviors or system dynamics of large-scale systems such as traffic flow \cite{lv2018lc}\cite{Hosseini2019traffic}, expression in genetic networks \cite{Maathuis2010}, reliability and failure prediction \cite{Zhu2013}, and prediction of atmospheric phenomena \cite{Charney2002}, to name just a few. These methods typically rely on model-free approaches such as neural networks which ``learn'' the underlying model from historical data to make predictions about future system states. These approaches have their roots in computer science-related disciplines, such as computer vision and natural language processing \cite{deng2013new}, using collected data to provide deeper insights into image and speech patterns and predict future system states \cite{sutskever2014sequence}. More recently, deep learning methods have experienced significant success in accurately predicting complex dynamics in large-scale multi-agent systems, such as predicting opponent's moves in the game of Go \cite{lecun2015deep}\cite{Silver2016}.

However, the behavior learned by data-driven approaches is limited by the amount of historical data that is available and may not capture the entire range of traffic flow dynamics. As opposed to these `model-free' approaches for predicting future traffic flow states, many modeling approaches, such as agent-based models \cite{Gehrke2008}\cite{Ando2006}, auto-regressive moving-average (ARMA) models \cite{Kamarianakis2012}\cite{Min2011} etc. can also be used to perform real-time or faster-than-real-time simulations to achieve a similar end. The key advantage of model-based approaches over data-driven approaches is the knowledge of model fidelity, i.e. the range of scenarios within which a specific model can be used is typically known. Among agent-based modeling approaches, cellular automata (CA)-based models have been known to capture traffic dynamics quite well and have been widely used in the transportation research community \cite{nagel1992cellular}.

In the presented work, the authors present a physics-based statistical mechanics model that has is closely related to the CA family of traffic flow models, but has the additional advantage that its parameters can be interpreted as physical quantities (e.g. interaction energy) \cite{jerath2014statistical}. The presented model is used to run simulations, create large training and test datasets not typically possible for traditional data-driven approaches, and subsequently examine the efficacy of deep learning methods in predicting future traffic flow states. The presented work uses Convolutional Neural Networks (CNNs), which offer advantages in terms of selecting good features, and Long Short-Term Memory (LSTM) Networks, which are proven to have the ability to learn sequential data \cite{lawrence1997face}\cite{sak2014long}. The presented deep learning method and CNN-LSTM architecture identifies patterns of traffic flow in the CA-based statistical mechanics model in both time and space, with the CNN being used to identify spatial patterns, and LSTM being used to determine temporal patterns, and is able to predict future traffic states to a reasonable degree.

The remainder of the paper is organized as follows: Section II discusses prior literature in the field of deep learning and its applications for traffic flow forecasting. Section III discusses the major components of the presented work, including the CA-based statistical mechanics model of traffic flow, and the  deep learning methods used to predict future traffic states via model-based algorithmically generated datasets that capture a broad range of traffic flow dynamics. Section IV summarizes the results and provides some perspectives on the upcoming future work.

\section{Literature review}

\subsection{Traffic Flow Prediction}
Traffic flow prediction plays a significant role in the development of Intelligent Transportation Systems, which relies on accurate traffic forecasting to support the control and management over a wide range of traffic conditions. Approaches would be changing according to desired outputs, among which state of traffic networks at some specific location in terms of vehicle density, average speed is most common. Extensively used for long term predictions, model-based approaches have been well developed for decades, including:
\begin{enumerate}
    \item \textbf{parametric approaches} like time-series-based predictions \cite{downs2010dynamic}, Kalman filtering \cite{whittaker1997tracking}\cite{okutani1984dynamic}, Gaussian Maximum Likelihood \cite{lin2001gaussian}, Cell Transmission Model \cite{daganzo1995cell}, shock-wave model \cite{richards1956shock}, Cellular Automata \cite{nagel1992cellular}, and
    \item \textbf{non-parametric approaches} like fuzzy neural approach \cite{yin2002urban}, sequential learning \cite{chen2001use} and Bayesian networks \cite{sun2006bayesian}.
\end{enumerate} 

Parametric approaches in traffic state prediction often aim to achieve accurate forecasts by incorporating road conditions such as network topology, speed limits, and traffic signals to varying degrees\cite{daganzo1995cell}\cite{lammer2008self}. Among these approaches, model order reduction (MOR) methods like proper orthogonal decomposition (POD), dynamic mode decomposition (DMD), and reduced order modeling (ROM) have gained attention for capturing the spatiotemporal dynamics inherent in dynamic systems\cite{yang2020observability}\cite{yang2018examining}, offering alternative perspectives on state prediction. For example, Yu\cite{yu2020low} demonstrated the efficacy of mode decomposition in characterizing temporally neighboring fragments of traffic flow via state transition matrices while maintaining satisfactory predictive performance. Similarly, research work\cite{xie2023novel} highlighted the suitability of Singular Value Decomposition for short-term traffic flow prediction, showing consistent results with the original data trends and offering insights into real-time traffic system characteristics for planning, control, and optimization. These studies underscore the potential of MOR approaches in enhancing the scalability and efficiency of traffic state prediction models, facilitating deployment in intelligent transportation systems (ITS)\cite{yang2022macroscopic}. However, while these approaches perform well under regular traffic variations, they may yield significant errors during emergent scenarios. To tackle such challenges, non-parametric approaches have garnered increasing attention, leading to the development of numerous models and algorithms to meet the growing demands for accurate traffic flow information. Additionally, a growing trend focuses on reducing model complexity and uncovering inherent patterns within the traffic data, even with limited observations \cite{yang2024multi}\cite{yang2020renormalization}.

Nowadays, data-based short term traffic predictions have been widely applied in industry to complement the active management of the road network through constant updates and assessments of road conditions \cite{pan2016traffic}. Apparently a shift to data-driven approaches is observable in recent research works as well. When traffic sensors were deployed on road networks with high time resolutions, information about traffic systems from various perspectives were contained in the `big data' produced by the sensors. This type of approach assumes the current traffic states would share certain patterns with historical databases. Challenges in how to achieve the transition from shallow correlation to deep correlations under the data have remained unsolved until the emergence of neural networks \cite{lv2014traffic}. 



\subsection{Applications of Deep Learning for Predicting Traffic Flow States}
Data-driven approaches have been widely implemented in traffic prediction as mentioned above, aided by the advancement in computational power that allows processing of ever larger datasets. Correspondingly, recent research on traffic prediction uses parameter regression, statistical estimations and deep neural network  to predict future states of a vehicle or traffic from historical data \cite{lv2014traffic}\cite{hofleitner2012learning}. Among these prior works, Convolutional Neural Network (CNN) have proven to perform well withh regards to predicting traffic flow and density across different scenarios by using time-space diagrams as inputs \cite{Hosseini2019traffic}. Compared with other neural networks, CNNs are better capable of extracting spatial features contained in images \cite{lawrence1997face}, and can perhaps be used to detect spatiotemporal patters in traffic flows as well. In addition to learning spatial interactions between neighboring sites in a CA-based traffic flow model, it is also important to recognize patterns in temporal sequences of data, i.e. time series of traffic flow as observed from a single spatial location or site \cite{lv2018lc}. For this purpose, the deep learning research community typically utilizes Recurrent Neural Networks (RNNs), which identify the functional dependence of the current hidden state $h_t$ at time step $t$ on a combination of the hidden state at the previous time step $h_{t-1}$ and the input at the current time step $x_t$. Thus, RNNs typically possess loops which allow information to be passed from one time step to the next \cite{mikolov2010recurrent}. A major drawback of RNNs is that the temporal range of contextual information is limited and the Back-Propagation through time does not work properly due to vanishing or exploding gradients. Consequently, recent works have begun to examine Long Short-Term Memory (LSTM) which has proven to be a more suitable approach when dealing with sequential data \cite{zhao2017lstm}.

Similar to the application of RNNs in natural language processing, the sequential data of spatial distributions in cellular automata models are correlated with neighboring sequences. In CA-based models, traffic states are updated recursively from the previous time step, and evolve along the temporal axis, and correlate well with the traffic dynamics in real life. LSTM neural networks, which contain loops to automatically preserve historical sequence information in its model structure, are able to capture the features relevant to time series within longer time span than traditional RNN \cite{hochreiter1997long}. 

However, most prior work directed towards predicting future traffic states from historical data is performed for fixed traffic scenarios and time periods. While these approaches provide reasonable accuracy within the narrow scope of application by using historical data, they do so by sacrificing robustness to environmental disturbances. Several key issues prove to be a detriment to widespread adoption of these purely data-driven approaches. First, these approaches lack portability of the `learned' behavior to different scenarios and prediction time windows. Second, the over reliance on historical data creates onerous data collection, storage and processing requirements, to capture even simple traffic flow patterns. Finally, these approaches disregard the rich trove of traffic flow models that have been built, examined, and validated by the transportation research community over the course of nearly a century. This paper seeks to remedy these issues by capturing the spatial interactions as well as temporal correlations in cellular automata-based statistical mechanics traffic flow model using deep neural networks, which combines the advantages of model-based and data-driven approaches to create a rich data set of known, calibrated and verifiable traffic flow dynamics which can be used to predict future states using machine learning methods in a wide range of traffic scenarios over nearly-arbitrary time windows.


\section{Deep learning in traffic prediction}

In contrast to prior works on prediction of traffic flow states which primarily rely on historical data, the authors propose to begin by incorporating traffic flow models into the deep learning methodology to generate vast training and test data sets that can be modified for arbitrary traffic scenarios. The ability to self-generate data through simulations also increases the likelihood of observing all possible traffic dynamics, including those that are not recorded in real-world data sets due to costs and observation limitations associated with data collection. The next section discusses the CA-based statistical mechanics traffic flow model which incorporates vehicle interaction physics, and is used to generate the training and test datasets for deep learning.

\subsection{CA-based Statistical Mechanics Model of Traffic Flow}
Traditionally, traffic flow is modeled at one of three spatial scales: microscopic models define dynamics of individual vehicles \cite{nagel1992cellular}\cite{gazis1961nonlinear}\cite{Bando1993}, macroscopic models describe traffic flow dynamics as a fluid \cite{Lighthill1955OnKW}\cite{Daganzo2016}, and mesoscopic models describe traffic at an intermediate scale as clusters \cite{jerath2014statistical}\cite{Kuhne2004}, cells \cite{daganzo1995cell}, or using position-velocity distributions \cite{Prigogine1957}. 
To generate training and test data for simulations, the choice of modeling approach is dictated by its ease of implementation and low computational overhead. Simulating traffic flow in discrete space and time, as in cellular automata based models provides these advantages. However, the parameters in CA-based traffic flow models do not have a clear physical interpretation, so the authors rely on a statistical mechanics modeling approach to simulate traffic flow. This approach has already been proven to faithfully replicate traffic flow characteristics \cite{jerath2014statistical}. Consequently, the \textbf{model is only briefly discussed below, and readers are directed to \cite{jerath2014statistical} for additional details}.


The CA-based statistical mechanics model draws inspiration from the Ising model, which is a prototypical model of a magnetic system and quite well known in the physics community. The Ising model of a magnetic system consists of atomic spins that can be in one of two states --- `up' spin (+1) and `down' spin (-1), corresponding to the magnetic moments of dipoles. Similar to the one-dimensional Ising model, traffic flow on a single-lane closed ring road can be modeled using $N$ cells with states $S_i \in \{1, 0\}$, where $(1)$ denotes a site occupied by a vehicle and $(0)$ denotes an empty site. 
In this model, site interactions extend beyond its nearest neighbor to a sequence of forward-looking neighborhood sites denoted by $\mathscr{N}(S_i)$, in accordance with the the driver's response in a real traffic \cite{jerath2014statistical}: 
\vspace{1em}

\begin{equation}\label{sec:forward_neighbor}
   \mathscr{N}(S_i)  = \{S_j: \: d(S_i, S_j)< d_l, \: j<i \}
\end{equation}

\vspace{1em}
\noindent where $d(S_i, S_j)$ is the distance between sites $S_i$ and $S_j$. The limit of site interactions is defined by the `look-ahead' distance $d_l$, and is set to $5$ in the simulations to indicate that vehicles can interact up to a distance of 5 sites or 25 m. The behavior of individual vehicles, i.e. the longitudinal vehicle dynamics are governed by the \textbf{Hamiltonian}, one of the most significant constructs in the statistical mechanics-inspired traffic model. The Hamiltonian is a measure of the energy of the traffic system, and is defined as follows:
\vspace{1em}

\begin{equation}\label{sec:Hamiltonian}
    H(S_i)  = -BS_i-\sum_{s_j \in \mathscr{N}(S_i)} K_{ij}S_iS_j
\end{equation}

\vspace{0.5em}
\noindent where $K_{ij}$ represents the  interaction coefficient between sites $S_i$ and $S_j$, and $B$ represents the external field coefficient. To account for the relative strength of interaction between sites that are further away, the site interaction coefficient about $K_{ij}$ varies with distance as follows:
\vspace{1em}

\begin{align}\label{sec:Interaction Coefficient}
\begin{aligned}
    K_{ij} = \frac{K_0}{(d(S_i, S_j))^2}
  \end{aligned}
\end{align}

\vspace{1em}
\noindent where $K_0$ is a constant, reflecting the base interaction strength. Detailed discussions about the choices made here, such as the functional form of $K_{ij}$ can be found in \cite{jerath2014statistical}. 

With the fundamentals of the model outlined, the Hamiltonian of the CA-based statistical mechanics model is used to determine the transition probability rate of a vehicle moving from one site to another. Since the traffic system being simulated is a closed ring-road (to avoid complicated analyses required for on- or off-ramps), the total number of vehicles is conserved. Consequently, the traffic system is modeled with periodic boundary conditions. The vehicles move forward according to `spin exchange dynamics', i.e. an occupied site can exchange its state with a vacant site immediately in front of it \cite{Alperovich2008}, transforming the state from $\sigma_t = \{S_1,S_2,...,S_{i-2}, 1, 0,S_{i+1},..., S_N\}$ at time $t$ to $\sigma_{t+1} = \{S_1,S_2,...,S_{i-2}, 0, 1,S_{i+1},..., S_N\}$ at time $t+1$. The `exchange' occurs between sites $S_{i-1}$ and $S_i$ and  represents the forward motion of an individual vehicle. The transition probability from one state to another is determined as $min(\phi, 1)$, where:

\begin{align}\label{sec: Transition probability}
    \phi & = -a_0 \: exp(-\beta H)
\end{align}

\vspace{1em}
\noindent where $\beta$ is a constant related to the vehicular density and $a_0$ is a pre-exponential factor \cite{jerath2014statistical}. Monte Carlo simulations are performed by recursively updating the states according to the Metropolis algorithm as shown in \textbf{Algorithm \ref{euclid}}. In this algorithm, the closed ring-road traffic system states are randomly initialized to various spatial configurations at time 0, and exchanges between an occupied site and adjacent forward empty site occur according to the Hamiltonian or energy of the states prior and after the exchange. If the energy decreases as a result of the exchange, then the new configuration is accepted, otherwise it is not. The pseudo-code for the algorithm is included below.

\begin{algorithm}[H] 
\caption{Metropolis Algorithm in the simulations of traffic flow }\label{euclid}
\begin{algorithmic}
\State{Initial condition: \:}{Vehicles are randomly distributed along an N-site ring road with a given vehicle density}
\While{$t< T_{end}$}
    \For{$i=1$ to $N$}
        \If {$S_i=1$ and $S_{i+1}=0$}
             \State{produce a random number $p \sim U(0,1)$}
             \State{calculate Hamiltonian $H$ of site $i$ }:
             \State{calculate transition probability $\phi=exp^{\{-\beta H\} }$}
             \If {$p<\phi$}
                \State{exchange site states of $S_i$ and $S_{i+1}$}
          \EndIf
        \EndIf
    \EndFor
\State $t \gets t+1$
\EndWhile\label{euclidendwhile}
\end{algorithmic}
\end{algorithm}

\subsection{Dataset Generation for Neural Network Training}
\begin{figure}[ht]
\centering
  \includegraphics[width=100mm]{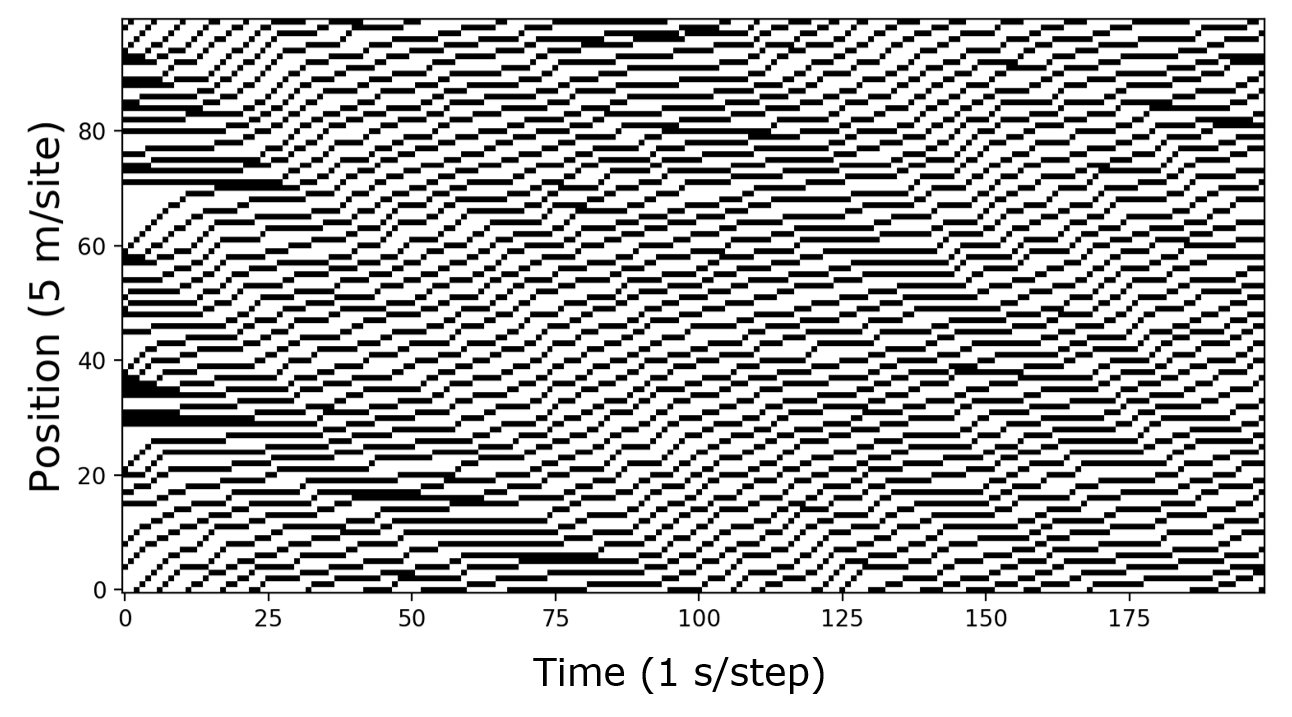}
	\caption{Time-space diagram of CA-based statistical mechanics model of traffic flow}
	\label{fig:Traffic flow}
\end{figure}

The CA-based statistical mechanics model of traffic flow was used to conduct Monte Carlo simulations and generate training and test data sets for a variety of traffic scenarios. One of these scenarios is shown in Figure \ref{fig:Traffic flow}: a 100-site traffic system which propagates at each time step along the horizontal axis for a total of 200 time steps, and in which vehicles are spatially distributed in the closed ring road along the vertical axis.

Prior to predicting future traffic states using deep learning, we need create a dataset for training and evaluation as well. In learning theory, it is desirable that the training sets adhere to the true population probability distributions, which results in a more accurate prediction for the future states. While such balanced and accurate data is undoubtedly favored in the training of neural network, it is infeasible to generate a dataset covering all possible scenarios in any deep learning approach.
Therefore, we have to devise methodologies to generate samples as a limited dataset, which is similar to the true population probability distribution.

From the analysis of \textbf{Hamiltonian} itself, interaction energy is the component directly influencing energy distributions of systems, while the other part of energy from external field would solely cause shifts in distributions. Hence, interaction energies in traffic systems with different sizes were compared to examine if any correlations existed among them. For systems with different number of sites, including $30, 60, 120, 240, 600$, initial conditions were randomly sampled for $3200$ times, to evaluate the interaction energy distributions and their relationship. 

\begin{figure}[h]
\centering
  \includegraphics[width=120mm]{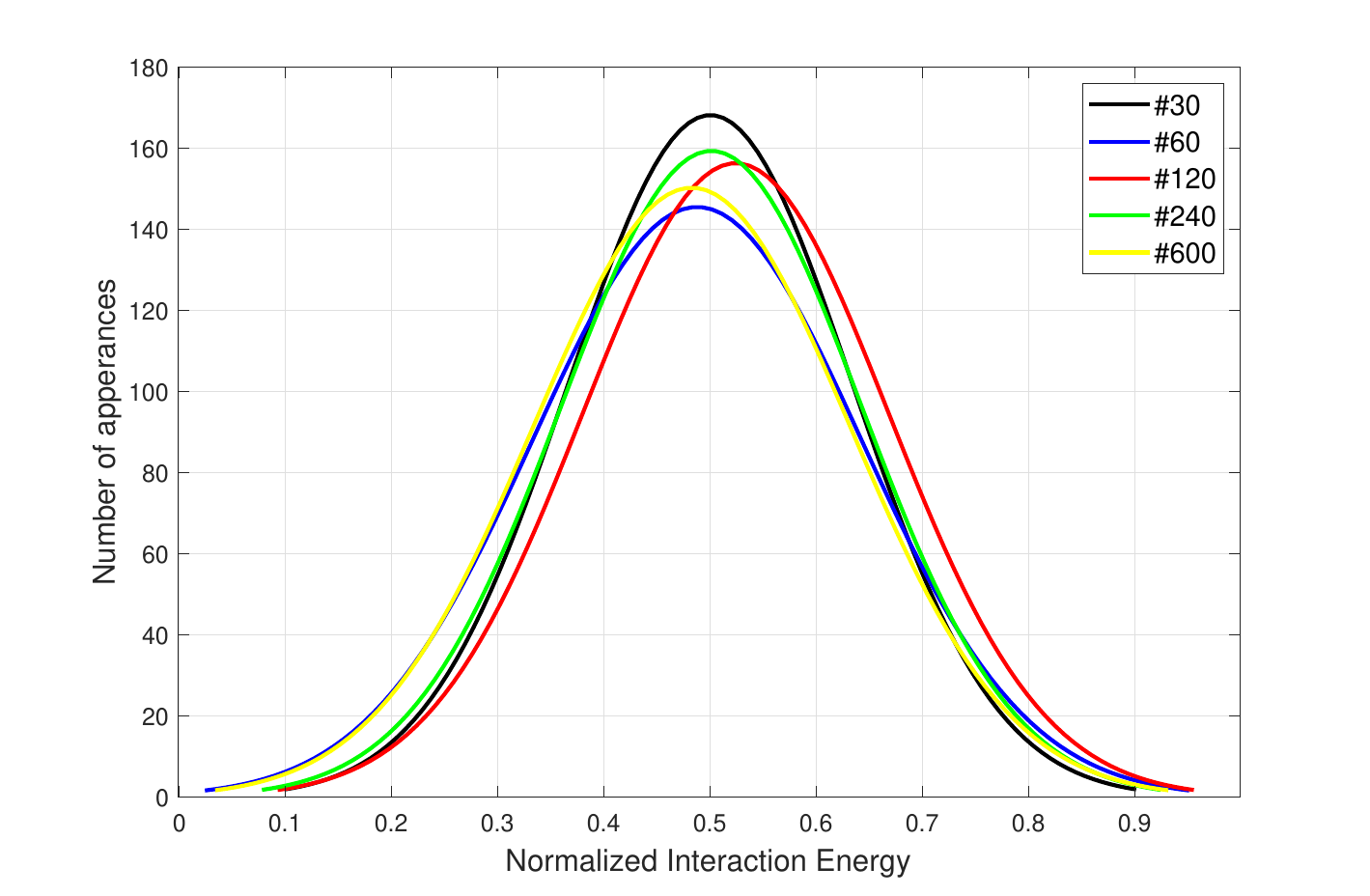}
	\caption{Normalized energy distributions of traffic systems with different spatial sizes}
	\label{fig:Energy}
\end{figure}

Clearly, Figure \ref{fig:Energy} reflects that the normalized energy distributions of periodic CA-based statistical mechanics models are similar to each other and scale-invariant. It means a high-quality dataset for a large traffic system could be sampled using the probability distribution function of a much smaller ring road traffic system. This is a \textbf{significant finding}, since the true probability distribution of systems with large sizes is much harder to acquire. Thus, the training and test dataset used in following sections on deep learning methods were obtained from a distribution derived from a reduced-size system (with much fewer sites). 
The simulated samples generated for training and testing in the following section are based on simulations with fewer sites and shorter time duration. Specifically, the training and test datasets are constructed such that the configurations of 50 sites sampled from $0-30$ time steps are assumed to be the inputs for the deep learning algorithm and the spatial distribution (i.e. the individual states of the 50 sites) at time step 31 is taken as the output. The next step involves using these data to train and test the deep learning neural network.

\subsection{CNN-LSTM Neural Network}
\begin{figure}[ht]
\centering
  \includegraphics[width=160mm]{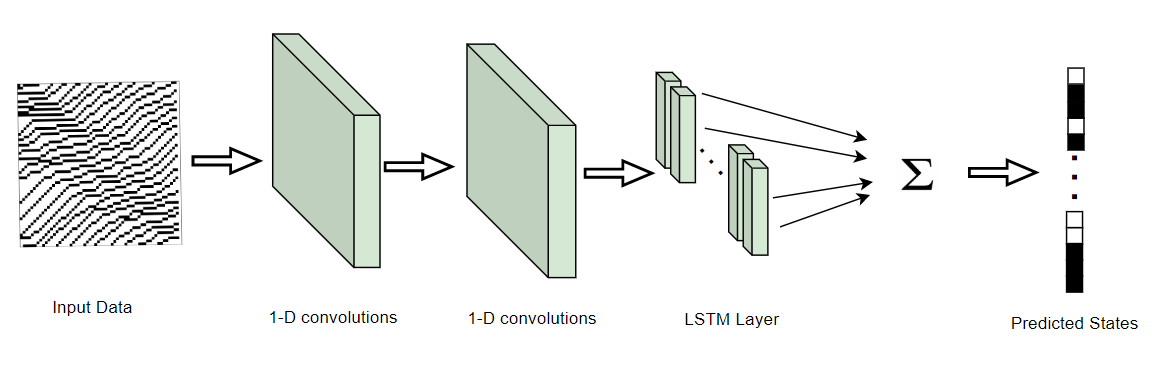}
	\caption{The framework of the CNN-LSTM model for traffic state prediction}
	\label{fig:CNN-LSTM}
\end{figure}

CNN-LSTM has demonstrated its significant predictive performance in Dimensional Sentiment Analysis \cite{wang2016dimensional}, and Conv-LSTM has been used to generate good predictions in traffic flow \cite{liu2017short}. 
At a specific time step, interactions between neighboring sites are recorded as features for the input state vector in spatial dimensions, which are assessed and identified by CNN layers. Similarly, at a specific spatial site but across several time steps, the distributions of CA site states are recorded as binary state sequences, which are identified via LSTM architecture. It is intuitive to think of the proposed deep learning architecture as a combination of two sub-models: the CNN model for feature extraction and LSTM model for interpreting the features across time steps. Based on this thought process, the deep learning neural network is constructed as shown in Figure \ref{fig:CNN-LSTM}.

The CNN submodel can be constructed using conventionally used deep learning structures. Realizing that a vehicle interacts with its neighboring sites in our CA-based statistical mechanics traffic model, two convolutional layers are connected locally after the initial fully connected layer. This architecture indicates that the output would be influenced only by its neighbors. Next, to avoid over-fitting of the trained models, a dropout layer is added with value 0.25 following the previous layer. During each time step, distributions of traffic states in space will be taken as a sequence by the neural network. Here, we added one LSTM layer to find the correlations between different adjacent sequences. 



\subsection{Assessment of the Neural Network Performance}
The CNN-LSTM-based neural network has been introduced in previous section, and data collected during simulations were used for parameter training. Here samples from 30-second simulations were taken as training inputs, and state of system at $31^{st}$ second as output. A total of 1000 such simulations were run to generate the dataset, with a split ratio of 0.2, i.e. with 80\% of the data being used to train the neural network and $20\%$ of the data being used to evaluate its performance. Since in our model, state of each site consists of two possible values, $1$ and $0$. \texttt{binary\_crossentropy} was used to calculate the loss at each epoch update as follows:
\vspace{1em}

\begin{align}\label{sec:binary crossentropy}
\begin{aligned}
    H(q) = -\frac{1}{N} \sum_{i=1}^{N}(y_i) \cdot log(p(y_i))+(1-y_i)\cdot log(1-p(y_i)) + \alpha (n_{vehicle} - \sum_{i=1}^{N} y_i) 
  \end{aligned}
\end{align}

\vspace{1em}
\noindent where $y_i$ is the state label (either 1 or 0), and $p(y)$ is the predicted probability of state being $1$. Here accuracy was defined based on the ratio of correct predictions in a sequence of traffic states. Convergence was guaranteed in the learning process with accuracy approaching 1.0 and loss proceeding towards zero, shown in Figure \ref{fig:accuracy loss record}: 

\begin{figure}[hbt!]
\centering
  \includegraphics[width=110mm]{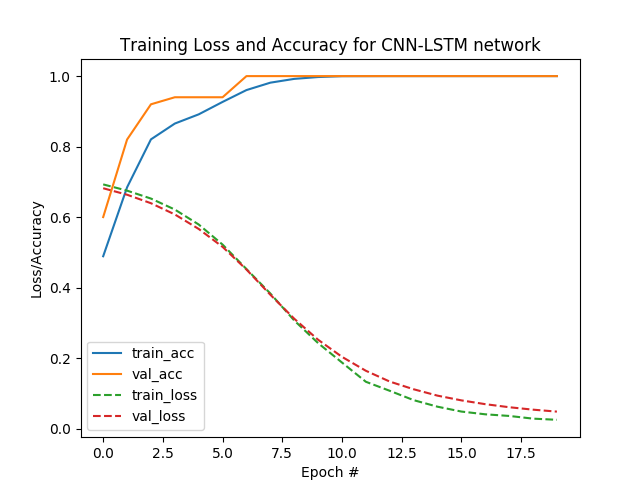}
	\caption{Loss and accuracy for traffic state prediction at one time step}
	\label{fig:accuracy loss record}
\end{figure}

\subsection{Traffic Flow State Prediction}
\begin{figure}[hbt!]
\centering
  \includegraphics[width=170mm]{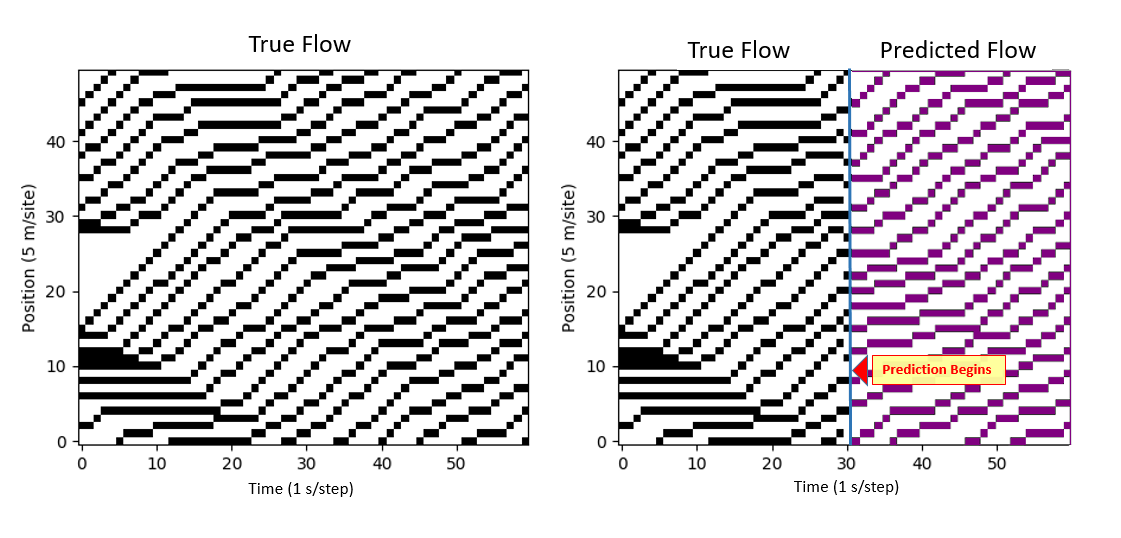}
	\caption{Comparisons of traffic flow between real simulations and CNN-LSTM based prediction}
	\label{fig:Traffic prediction}
\end{figure}

The CNN-LSTM based neural network was trained using $1000$ images (corresponding to the 1000 simulations of the traffic flow model) with $50 \times 30$ pixels. Each image was sampled from energy based distribution and contained the traffic flow information for 50-site traffic systems over a duration of $30$ seconds. The size of dataset is extremely small compared with the total number of possible traffic state configurations (which is $\approx 2^{14}$ covering all possible initial conditions), given traffic density is $0.5$. To visualize the performance of the trained neural network, we compare it against the true data (i.e. the simulated traffic flow). A series of states were predicted iteratively using the past consecutive $30$ time steps, so that states predicted at each step would be added as the last vector of new inputs for future predictions. In other words, states between time intervals $1\sim30$, $2\sim31$,...., $30\sim 59$ were progressively taken as inputs and states at time $31$, $32$,...., $60$ were the corresponding outputs. Accordingly, spatial-temporal diagram was gradually expanded from $30$ seconds to $60$ seconds. Comparing the two plots in Figure \ref{fig:Traffic prediction}, the left figure is the true plot of traffic flow with 60-second span and the right figure depicts the predicted states obtained using the trained neural networks when states of only first 30 seconds were given.

\section{Results and conclusions}

In this paper, deep neural networks were trained to forecast traffic flow using a CA-based statistical mechanics traffic flow model to capture a large dynamical space using a small dataset. Results from Monte Carlo simulations in Figure \ref{fig:Energy} showed that the normalized energy distributions are close to each other for cellular automata models despite their different sizes. Hence, a balanced, representative dataset could be generated for a large traffic system by sampling the distribution of a much smaller size system.

The CNN-LSTM architecture was used to capture the correlation both in spatial and temporal space. The test samples provide proof about the convergence as well as good accuracy in one step prediction. In this paper, we utilized the neural network to do flow prediction step by step as Markov process, and 30 iterations were conducted to get a time-space diagram. While the results seem to indicate some resemblance with simulated traffic flow, additional research needs to be performed to improve the accuracy of the neural network by using larger datasets for training, both in terms of the number of simulations and the size of the traffic system. Partial reasons for incorrect predictions in some sites are the trade-off among the complexity and accuracy. For example, a small dataset may be unable to train a complicated neural network while keeping a relatively high accuracy in prediction, since this typically requires networks with deep layers and more parameters. Additionally, the CNN-LSTM architecture also results in negative influence on performance due to error expansions in recursive predictions. Results in this paper provide new ideas on data sampling from the perspective of energy to guarantee high quality predictions. Moreover, it also provides proof that phenomena that can be modeled using statistical mechanics approaches may be learned from past data and make accurate predictions about the future.

\section{Author Contribution Statement}

The authors confirm contribution to the paper as follows: study conception and design: Z. Yang, K. Jerath; analysis and interpretation of results: Z. Yang, K. Jerath; draft manuscript preparation: Z. Yang, K. Jerath. All authors reviewed the results and approved the final version of the manuscript.

\bibliographystyle{unsrt}  

\bibliography{references}

\end{document}